\documentclass[10pt,twocolumn,letterpaper]{article}

\usepackage{iccv}

\usepackage{times}
\usepackage{relsize}
\usepackage[T1]{fontenc} 
\usepackage[latin1]{inputenc}
\usepackage[english]{babel}

\usepackage{afterpage}
\usepackage{epsfig}
\usepackage{graphicx}
\usepackage{wrapfig}
\usepackage[belowskip=-2pt,aboveskip=3pt,font=small]{caption}
\usepackage[belowskip=0pt,aboveskip=3pt,font=small]{subcaption}
\usepackage{amsmath, amsthm, amssymb}
\usepackage{textcomp}
\usepackage{stmaryrd}
\usepackage{upgreek}
\usepackage{bm}
\usepackage{cases}
\usepackage{mathtools}
\usepackage{arydshln}
\usepackage{multirow}

\usepackage{cite}
\usepackage[pagebackref=true,breaklinks=true,letterpaper=true,colorlinks,bookmarks=false,citecolor=green]{hyperref}
\usepackage{bibspacing}
\usepackage{algorithm}
\usepackage{algpseudocode}

\usepackage{multirow}
\usepackage{rotating}
\usepackage{booktabs}

\usepackage{enumitem}
\usepackage[olditem,oldenum]{paralist}

\usepackage{alltt}
\usepackage{listings}

\usepackage{mysymbols}

\usepackage{url}
\usepackage{xspace}
\usepackage{comment}
\usepackage{color}
\usepackage{xcolor}
\usepackage{afterpage}
\usepackage{pdfpages}
\usepackage{framed}
\usepackage{fancybox}
\usepackage{cuted}
\usepackage{caption}

\belowdisplayskip \abovedisplayskip

\newlength{\sectionReduceTop}
\newlength{\sectionReduceBot}
\newlength{\subsectionReduceTop}
\newlength{\subsectionReduceBot}
\newlength{\abstractReduceTop}
\newlength{\abstractReduceBot}
\newlength{\captionReduceTop}
\newlength{\captionReduceBot}
\newlength{\subsubsectionReduceTop}
\newlength{\subsubsectionReduceBot}

\newlength{\eqnReduceTop}
\newlength{\eqnReduceBot}

\newlength{\horSkip}
\newlength{\verSkip}

\newlength{\figureHeight}
\newcommand{\haty}{\hat{y}}

\newcommand{\Qbot}{\textsc{Q-bot}\xspace}
\newcommand{\Abot}{\textsc{A-bot}\xspace}

\newcommand{\fq}{\texttt{Frozen-Q}}
\newcommand{\fqm}{\texttt{Frozen-Q-multi}}
\newcommand{\fa}{\texttt{Frozen-A}}

\newcommand{\slp}{\texttt{SL-pretrained}\xspace}
\newcommand{\rl}{\texttt{RL-full-QAf}\xspace}

\newcommand{\myquote}[1]{\emph{`#1'}}

\newcommand{\reffig}[1]{Fig.~\ref{#1}}
\newcommand{\refsec}[1]{Sec.~\ref{#1}}
\newcommand{\reftab}[1]{Tab.~\ref{#1}}

\newcommand{\train}{\texttt{train}\xspace}

\newcommand{\test}{\texttt{test}\xspace}

\newcommand{\ad}[1]{\textcolor{teal}{#1}}

\newcommand{\sml}[1]{\textcolor{black}{#1}}

\iccvfinalcopy 

\def\iccvPaperID{175} 

\usepackage{flushend}

\begin{document}

\title{Learning Cooperative Visual Dialog Agents with Deep Reinforcement Learning}

 \author{
Abhishek Das$^1$\thanks{The first two authors (AD, SK) contributed equally.}, \,\,
Satwik Kottur$^2$\footnotemark[1], \,\,
Jos\'e M.F. Moura$^2$, \,\,
Stefan Lee$^3$, \,\,
Dhruv Batra$^1$ \\
$^1$Georgia Institute of Technology, $^2$Carnegie Mellon University, $^3$Virginia Tech\\
\tt\normalsize \href{http://visualdialog.org}{visualdialog.org}
 }

\maketitle

\begin{abstract}

We introduce the first goal-driven training for visual question answering and dialog agents.
Specifically, we pose a
cooperative `image guessing' game between two agents  -- \Qbot and \Abot --
who communicate in natural language dialog so that \Qbot can select an unseen image
from a lineup of images.
We use deep reinforcement learning (RL) to learn the policies of these agents 
end-to-end -- from pixels to multi-agent multi-round dialog to game reward.

We demonstrate two experimental results.

First, as a `sanity check' demonstration of pure RL (from scratch),
we show results on a synthetic world, where the agents communicate
in ungrounded vocabulary, \ie, symbols with no pre-specified meanings (X, Y, Z).
We find that two bots \emph{invent their own communication protocol}
and start using certain symbols to ask/answer about certain visual attributes (shape/color/style).
Thus, we demonstrate the \emph{emergence of grounded language and communication}
among `visual' dialog agents with no human supervision.

Second, we conduct large-scale real-image experiments on the VisDial dataset~\cite{visdial}, where
we pretrain with supervised dialog data and
show that the RL `fine-tuned' agents significantly outperform SL agents.
Interestingly, the RL \Qbot learns to ask questions that \Abot is good at,
ultimately resulting in more informative dialog and a better team.

\end{abstract}

\vspace{-7pt}
\section{Introduction}
\label{sec:intro}

The focus of this paper is
visually-grounded conversational artificial intelligence (AI). Specifically, we would like to develop 
agents that can `see' (\ie, understand the contents of an image) and `communicate' that understanding in 
natural language (\ie, hold a dialog involving questions and answers about that image).
We believe the next generation of intelligent systems will need to posses this ability to hold a dialog
about visual content for a variety of applications: \eg, 
helping visually impaired users understand their surroundings~\cite{bigham_uist10} or social media content~\cite{fb_blog16}  (\myquote{Who is in the photo? Dave. What is he doing?}), 
enabling analysts to sift through large quantities of surveillance data 
(\myquote{Did anyone enter the vault in the last month? Yes, there are 103 recorded instances. Did any of them pick something up?}), 
and enabling users to interact naturally with intelligent assistants (either embodied as a robot or not) 
(\myquote{Did I leave my phone on my desk? Yes, it's here. Did I miss any calls?}).

\begin{figure}[t]
  \centering
	\includegraphics[width=1\columnwidth]{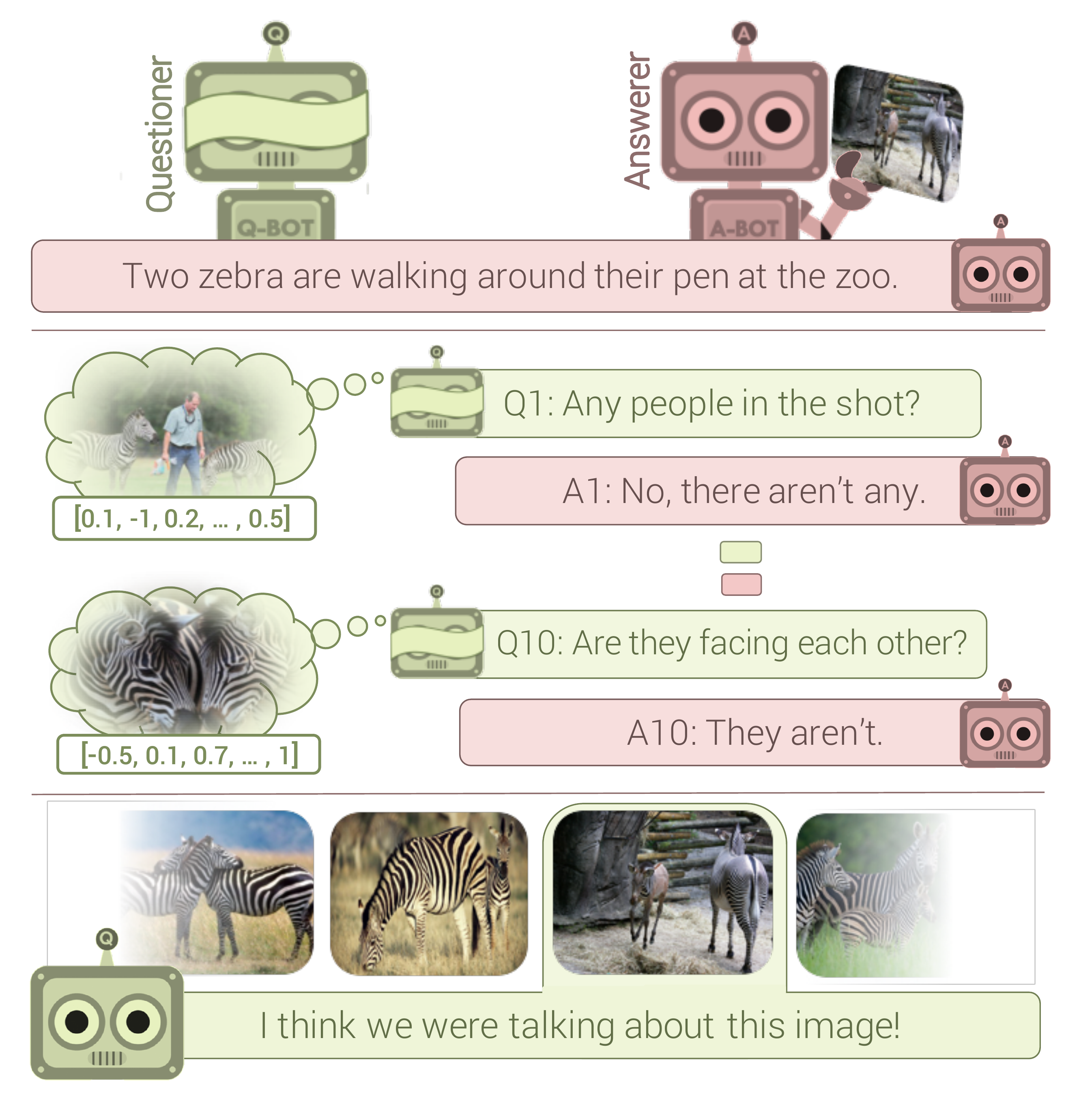}
  \caption{We propose a cooperative image guessing game between two agents -- \Qbot and \Abot -- who communicate 
  through a natural language dialog so that \Qbot can select a particular unseen image from a lineup. 
  We model these agents as deep neural networks and train them end-to-end with reinforcement learning.}
  \label{fig:teaser}
\end{figure}

Despite rapid progress at the intersection of vision and language, in particular, in image/video captioning~\cite{vinyals_cvpr15,chen_cvpr15,xu_icml15,johnson_cvpr16,venugopalan_iccv15,venugopalan_naacl15} 
and question answering~\cite{antol_iccv15,ren_nips15,malinowski_iccv15,tapaswi_cvpr16,tu_mm14}, 
it is clear we are far from this grand goal of a visual dialog agent. 

Two recent works~\cite{visdial,vries_cvpr17} have proposed studying this task of visually-grounded dialog. 
Perhaps somewhat counterintuitively, both these works treat dialog as a \emph{static} supervised learning problem, 
rather than an \emph{interactive} agent learning problem that it naturally is. 
Specifically, both works~\cite{visdial,vries_cvpr17} first collect a dataset of human-human dialog, 
\ie, a sequence of question-answer pairs about an image $(q_1, a_1), \ldots, (q_{T}, a_{T})$.   
Next, a machine (a deep neural network) is provided with the image $I$, the human dialog recorded till round $t-1$, 
$(q_1, a_1), \ldots, (q_{t-1}, a_{t-1})$, 
the follow-up question $q_t$, and is supervised to generate the human response $a_t$. 
Essentially, at each round $t$, the machine is artificially `injected' into the conversation between two humans 
and asked to answer the question $q_t$; 
but the machine's answer $\hat{a}_t$ is thrown away, because at the next round $t+1$, 
the machine is again provided with the `ground-truth' human-human dialog that includes the human response 
$a_t$ and not the machine response $\hat{a}_t$. 
Thus, the machine is \emph{never allowed to steer the conversation} because that would take the dialog out of the dataset, making it non-evaluable. 

In this paper, we generalize the task of Visual Dialog beyond the necessary first stage of 
supervised learning -- by posing it as a cooperative `image guessing' game 
between two dialog agents. We use deep reinforcement learning (RL) to 
learn the policies of these agents end-to-end -- 
from pixels to multi-agent multi-round dialog to the game reward. 


Our setup is illustrated in \figref{fig:teaser}. We formulate a game between 
a questioner bot (\Qbot) and an answerer bot (\Abot). 
\Qbot is shown a 1-sentence description (a caption) of an unseen image, 
and is allowed to communicate in natural language 
(discrete symbols) with the answering bot (\Abot), who is shown the image. 
The objective of this fully-cooperative game is for \Qbot to build a mental model of the unseen image 
purely from the natural language dialog, 
and then retrieve that image from a lineup of images.

Notice that this is a challenging game. \Qbot must ground the words mentioned in the provided caption 
(\myquote{Two zebra are walking around their pen at the zoo.}), 
estimate which images from the provided pool contain this content 
(there will typically be many such images since captions describe only the salient entities), 
and ask follow-up questions (\myquote{Any people in the shot? 
Are there clouds in the sky? Are they facing each other?}) 
that help it identify the correct image. 

Analogously, \Abot must build a mental model of what \Qbot understands, 
and answer questions (\myquote{No, there aren't any. I can't see the sky. They aren't.})  
in a precise enough way to allow discrimination between 
similar images from a pool (that \Abot does not have access to) while being 
concise enough to not confuse the imperfect \Qbot.

At every round of dialog, \Qbot listens to the answer provided by \Abot, updates its beliefs, 
and makes a prediction about the visual representation of the unseen image 
(specifically, the fc7 vector of $I$), and receives a reward from the environment based on how 
close \Qbot's prediction is to the true fc7 representation of $I$. 
The goal of \Qbot and \Abot is to communicate to maximize this reward. 
One critical issue is that both the agents 
are imperfect and noisy -- both `forget' things in the past, sometimes repeat themselves, 
may not stay consistent in their responses, 
\Abot does not have access to an external knowledge-base so it cannot answer all questions, \etc. 
Thus, to succeed at the task, they must learn to play to each other's strengths.

An important question to ask is -- why force the two agents to communicate in discrete symbols 
(English words) as opposed to continuous vectors? 
The reason is twofold. First, discrete symbols and natural language is interpretable. 
By forcing the two agents to communicate and understand natural language, 
we ensure that humans can not only inspect the conversation logs between two agents, 
but more importantly, communicate with them. 
After the two bots are trained, we can pair a human questioner with \Abot to accomplish 
the goals of visual dialog (aiding visually/situationally impaired users), 
and pair a human answerer with \Qbot to play a visual 20-questions game. 
The second reason to communicate in discrete symbols is to prevent cheating -- 
if \Qbot and \Abot are allowed to exchange continuous vectors, 
then the trivial solution is for \Abot to ignore \Qbot's question and directly convey the fc7 vector for $I$, 
allowing \Qbot to make a perfect prediction.
In essence, discrete natural language is an interpretable low-dimensional ``bottleneck'' 
layer between these two agents.

\textbf{Contributions.} 
We introduce a novel goal-driven training for visual question answering and dialog agents. 
Despite significant popular interest in VQA (over 200 works citing \cite{antol_iccv15} since 2015), 
all previous approaches have been based on supervised learning, making this the first instance of 
\emph{goal-driven} training for visual question answering / dialog. 

We demonstrate two experimental results. 

First, as a `sanity check' demonstration of pure RL (from scratch), 
we show results on a diagnostic task where perception is perfect -- 
a synthetic world with `images' containing a single object defined by three attributes (shape/color/style).
In this synthetic world, for \Qbot to identify an image, it must learn about these attributes. 
The two bots communicate via an ungrounded vocabulary, 
\ie, symbols with no pre-specified human-interpretable meanings (`X', `Y', `1', `2'). 
When trained end-to-end with RL on this task, 
we find that the two bots \emph{invent their own communication protocol} 
-- \Qbot starts using certain symbols to query for specific attributes (`X' for color), 
and \Abot starts responding with specific symbols indicating the value of that attribute 
(`1' for red). 
Essentially, we demonstrate the \emph{automatic emergence of grounded language and communication} 
among `visual' dialog agents with no human supervision!

Second, we conduct large-scale real-image experiments on the VisDial dataset~\cite{visdial}.  
With imperfect perception on real images, discovering a human-interpretable language 
and communication strategy from scratch is both tremendously difficult 
and an unnecessary re-invention of English. 
Thus, we pretrain with supervised dialog data in VisDial before `fine tuning' with RL;  
this alleviates a number of challenges in making deep RL converge to something meaningful. 
We show that these RL fine-tuned bots significantly outperform the supervised bots. 
Most interestingly, while the supervised \Qbot attempts to mimic how humans ask questions, 
the RL trained \Qbot \emph{shifts strategies} and asks questions that the \Abot is better at answering, 
ultimately resulting in more informative dialog and a better team. 
\section{Related Work}
\label{sec:related}

\textbf{Vision and Language.}
A number of problems at the intersection of vision and language have recently gained prominence,
\eg, image captioning~\cite{fang_cvpr15,vinyals_cvpr15,karpathy_cvpr15,donahue_cvpr15},
and visual question answering (VQA)~\cite{malinowski_nips14,antol_iccv15,ren_nips15,gao_nips15,malinowski_iccv15}.
Most related to this paper are two recent works
on visually-grounded dialog~\cite{visdial,vries_cvpr17}.
Das~\etal~\cite{visdial} proposed the task of Visual Dialog, collected the VisDial dataset
by pairing two subjects on Amazon Mechanical Turk to chat about an image
(with assigned roles of `Questioner' and `Answerer'),
and trained neural visual dialog answering models.
De Vries~\etal~\cite{vries_cvpr17} extended the Referit game~\cite{kazemzadeh_emnlp14} to a
`GuessWhat' game, where one person asks questions about an image to guess which object has been `selected',
and the second person answers questions in `yes'/`no'/NA (natural language answers are disallowed).
One disadvantage of GuessWhat is that it requires bounding box annotations for objects; our image
guessing game does not need any such annotations and thus an unlimited number of game plays may be simulated.
Moreover, as described in \refsec{sec:intro}, both these works unnaturally treat dialog
as a static supervised learning problem.
Although both datasets contain thousands of human dialogs,
they still only represent an incredibly sparse sample of the vast space of visually-grounded questions
and answers. Training robust, visually-grounded dialog agents via supervised techniques is still a challenging task.


In our work, we take inspiration from the AlphaGo~\cite{alphago} approach
of supervision from human-expert games and reinforcement learning from self-play.
Similarly, we perform supervised pretraining on human dialog data and fine-tune
in an end-to-end goal-driven manner with deep RL.
\textbf{20 Questions and Lewis Signaling Game.}
Our proposed image-guessing game is naturally the visual analog of the popular 20-questions game.
More formally, it is a generalization of the Lewis Signaling (LS)\cite{lewis2008convention} game, widely studied in economics and game theory.
LS is a cooperative game between two players -- a \emph{sender} and a \emph{receiver}. In the classical setting,
the world can be in a number of finite discrete states $\{1, 2, \ldots, N\}$, which is known to the sender
but not the receiver. 
The sender can send one of $N$ discrete symbols/signals to the receiver, who upon receiving the signal
must take one of $N$ discrete actions.
The game is perfectly cooperative, and one simple (though not unique) Nash Equilibrium is the `identity mapping', where the sender encodes each world state
with a bijective signal, and similarly the receiver has a bijective mapping from a signal to an action.

Our proposed `image guessing' game is a generalization of LS with \Qbot being the receiver and \Abot the sender.
However, in our proposed game, the receiver (\Qbot) is not passive. It actively solicits information by asking questions. Moreover, the signaling process is not
`single shot', but proceeds over multiple rounds of conversation.

\textbf{Text-only or Classical Dialog.}
Li~\etal~\cite{li_emnlp16} have proposed using RL for training dialog systems. However, they hand-define what a
`good' utterance/dialog looks like (non-repetition, coherence, continuity, \etc).
In contrast, taking a cue from adversarial learning~\cite{goodfellow_nips14,li2017adversarial},
we set up a cooperative game between two agents, such that we do not need to hand-define what a `good'
dialog looks like -- a `good' dialog is one that leads to a successful image-guessing play.



\textbf{Emergence of Language.}
There is a long history of work on language emergence in multi-agent systems~\cite{nolfi_springer09}.
The more recent resurgence has focused on
deep RL~\cite{foerster_nips16,lazaridou_iclr17,havrylov_iclr17,mordatch_arxiv17}.
The high-level ideas of these concurrent works are similar to our synthetic experiments. 
For our large-scale real-image results, we do not want our bots to invent their own uninterpretable language
and use pretraining on VisDial~\cite{visdial} to achieve `alignment' with English.

\section{Cooperative Image Guessing Game: \\ {\small In Full Generality and a Specific Instantiation}} 
\label{sec:game}


\textbf{Players and Roles.}
The game involves two collaborative agents -- a questioner bot (\Qbot) and an answerer bot (\Abot) --
with an information asymmetry.
\Abot sees an image $I$, \Qbot does not.
\Qbot is primed with a 1-sentence description 
$c$ of the unseen image and asks
`questions' (sequence of discrete symbols over a vocabulary $V$), which \Abot answers
with another sequence of symbols. The communication occurs for a fixed number
of rounds.

\textbf{Game Objective in General.}
At each round, in addition to communicating,
\Qbot must provide a `description' $\haty$ of the unknown image $I$
based only on the dialog history and both players receive a reward from the environment
inversely proportional to the error in this description
under some metric $\ell(\haty, y^{gt})$. We note that this is a general setting
where the `description' $\haty$ can take on varying levels of specificity -- from image
embeddings (or fc7 vectors of $I$) to textual descriptions to pixel-level image generations.

\textbf{Specific Instantiation.}
In our experiments, we focus on the setting where \Qbot is tasked with estimating a vector
embedding of the image $I$. Given some feature extractor (\ie, a pretrained CNN model, say VGG-16),
no human annotation is required to produce the target `description' $\haty^{gt}$
(simply forward-prop the image through the CNN).
Reward/error can be measured by simple Euclidean distance, and any image may be
used as the visual grounding for a dialog. Thus, an unlimited number of `game plays' may be simulated.

\section{Reinforcement Learning for Dialog Agents}
\label{sec:model}


In this section, we formalize the training of two visual dialog agents (\Qbot and \Abot) with Reinforcement Learning (RL) 
-- describing formally the \emph{action}, \emph{state}, \emph{environment}, \emph{reward}, \emph{policy},  
and training procedure.
We begin by noting that although there are two agents (\Qbot, \Abot), since the game is perfectly cooperative, 
we can without loss of generality view this as a single-agent RL setup where the single 
``meta-agent'' comprises of two ``constituent agents'' communicating via a natural language bottleneck layer. 

\begin{figure*}[t]
  \centering
\includegraphics[width=1\textwidth]{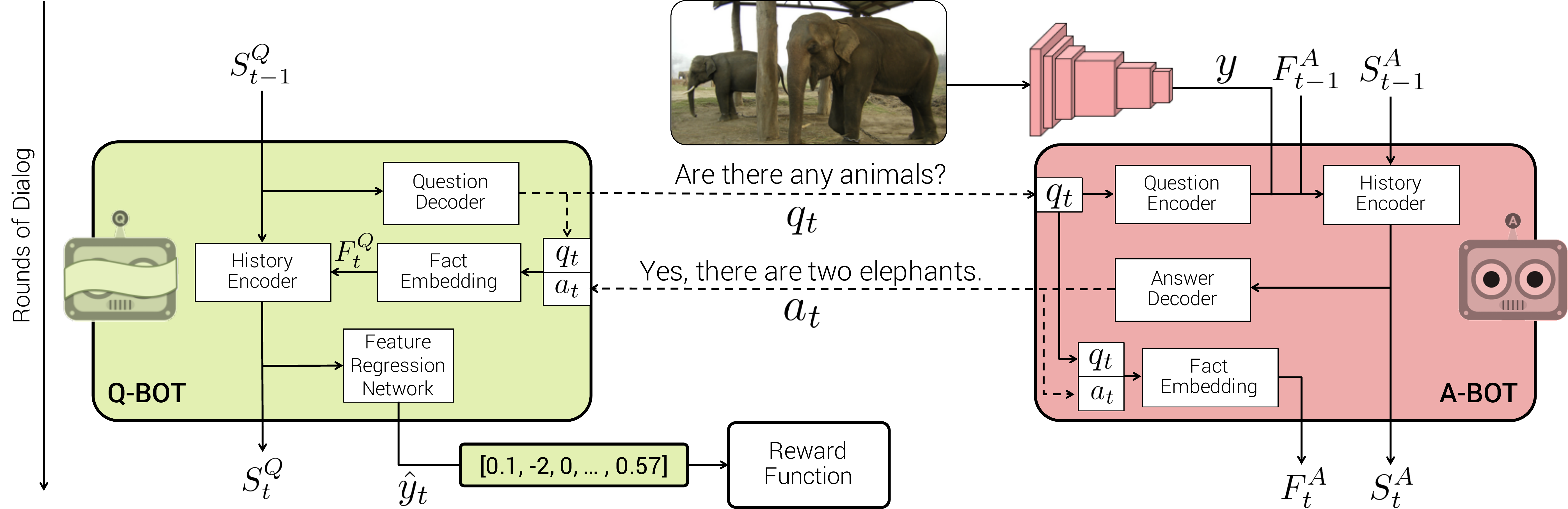}\vspace{5pt}
  \caption{Policy networks for \Qbot and \Abot. 
  At each round $t$ of dialog, 
  (1) \Qbot generates a question $q_t$ from its question decoder conditioned on its state encoding $S^Q_{t-1}$, 
  (2) \Abot encodes $q_t$, updates its state encoding $S^A_t$, and generates an answer $a_t$, 
  (3) both encode the completed exchange as $F^Q_t$ and $F^A_t$, and 
  (4) \Qbot updates its state to $S^Q_t$, predicts an image representation $\haty_t$, and receives a reward.   
  }
  \label{fig:models}
\end{figure*}

\textbf{Action.}
Both agents share a common action space consisting of all possible output sequences under a token vocabulary $V$. 
This action space is discrete and in principle, infinitely-large since arbitrary length sequences $q_t, a_t$ 
may be produced and the dialog may go on forever. 
In our synthetic experiment, the two agents are given different vocabularies to coax a certain 
behavior to emerge (details in \refsec{sec:exp_toy}). In our VisDial experiments, the two agents share a common vocabulary 
of English tokens. In addition, at each round of the dialog $t$, \Qbot also predicts $\haty_t$, its current guess about 
the visual representation of the unseen image. This component of \Qbot's action space is continuous. 

\textbf{State.}
Since there is information asymmetry (\Abot can see the image $I$, \Qbot cannot), each agent has its own 
observed state. 
For a dialog grounded in image $I$ with caption $c$, 
the state of \Qbot at round $t$ is the caption and dialog history so far 
$s^Q_t = \left[c, q_1, a_1, \dots, q_{t-1}, a_{t-1}\right]$, 
and the state of \Abot also includes the image 
$s^A_t = \left[I, c, q_1, a_1, \dots, q_{t-1}, a_{t-1}, q_t\right]$. 

\textbf{Policy.}
We model \Qbot and \Abot operating under stochastic policies $\pi_Q(q_t \mid s^Q_t; \theta_Q)$ 
and $\pi_A(a_t \mid s^A_t; \theta_A)$, such that questions and answers may be sampled from these policies 
conditioned on the dialog/state history. 
These policies will be learned by two separate deep neural networks 
parameterized by $\theta_Q$ and $\theta_A$. 
In addition, \Qbot includes a feature regression network $f(\cdot)$ that produces 
an image representation prediction \emph{after listening to the answer at round $t$}, 
\ie, $\haty_t = f(s^Q_t, q_t, a_t; \theta_f) = f(s^Q_{t+1}; \theta_f)$. 
Thus, the goal of policy learning is to estimate the parameters $\theta_Q, \theta_A, \theta_f$.


\textbf{Environment and Reward.}
The environment is the image $I$ upon which the dialog is grounded. 
Since this is a purely cooperative setting, both agents receive the same reward. 
Let $\ell(\cdot,\cdot)$ be a distance metric on image representations 
(Euclidean distance in our experiments). 
At each round $t$, we define the reward for a state-action pair as: \\[-22pt]

\begin{center}
{
\begin{minipage}{\columnwidth}
\begin{align}
r_t \Big( \underbrace{s^Q_t}_{\text{state}}, \underbrace{(q_t, a_t, y_t)}_{\text{action}} \Big) 
= \underbrace{\ell\left(\haty_{t-1}, y^{gt}\right)}_{\text{distance at $t$-1}} 
- \underbrace{\ell\left(\haty_t, y^{gt}\right)}_{\text{distance at $t$}}
\end{align}
\end{minipage}}
\end{center}
\vspace{-3pt}

\noindent\ie, the \emph{change in distance} to the true representation before and after a round of dialog. 
In this way, we consider a question-answer pair to be low quality (\ie, have a negative reward) 
if it leads the questioner to make a \emph{worse} estimate of the target image representation than if the dialog had ended.

Note that the total reward summed over all time steps of a dialog is a function 
of only the initial and final states due to the cancellation of intermediate terms, \ie, \\[-20pt]
\begin{center}
{
\begin{minipage}{\columnwidth}
\begin{align}
\sum_{t=1}^T r_t \Big(s^Q_t, (q_t, a_t, y_t)) \Big) = 
\underbrace{\ell\left(\haty_0, y^{gt}\right) - \ell\left(\haty_T, y^{gt}\right)}_{\text{overall improvement due to dialog}}
\end{align}
\end{minipage}}
\end{center}
%
\noindent This is again intuitive -- 
`How much do the feature predictions of \Qbot improve due to the dialog?' 
The details of policy learning are described in \refsec{sec:joint}, but before that, let us describe the 
inner working of the two agents.

\subsection{Policy Networks for \Qbot and \Abot}

\reffig{fig:models} shows an overview of our policy networks for \Qbot and \Abot and their 
interaction within a single round of dialog. 
Both the agent policies are modeled via Hierarchical Recurrent Encoder-Decoder neural networks, 
which have recently been proposed for dialog modeling~\cite{serban_arxiv16,serban_aaai16,visdial}. 

\textbf{\Qbot} consists of the following four components: 

\begin{compactenum}[\hspace{0pt}-]

\vspace{3pt}
\item \textbf{Fact Encoder}: \Qbot asks a question $q_t$: \myquote{Are there any animals?} and 
receives an answer $a_t$: \myquote{Yes, there are two elephants.}. \Qbot treats this concatenated 
$(q_t, a_t)$-pair as a `fact' it now knows about the unseen image. 
The fact encoder is an LSTM whose final hidden state $F^Q_t \in \RF^{512}$ 
is used as an embedding of $(q_t, a_t)$.  

\vspace{4pt}
\item \textbf{State/History Encoder} 
is an LSTM that takes the encoded fact $F^Q_t$ at each time step to produce an encoding of the prior 
dialog including time $t$ as $S^Q_t \in \RF^{512}$. 
Notice that this results in a two-level hierarchical encoding of the dialog  $(q_t, a_t) \rightarrow F^Q_t$ 
and $(F^Q_1, \ldots, F^Q_t) \rightarrow S^Q_t$. 

\vspace{4pt}
\item \textbf{Question Decoder} 
is an LSTM that takes the state/history encoding from the previous round $S^Q_{t-1}$ 
and generates question $q_t$ by sequentially sampling words. 

\vspace{4pt}
\item \textbf{Feature Regression Network} 
$f(\cdot)$ is a single fully-connected layer 
that produces an image representation prediction $\haty_t$ from the current 
encoded state $\haty_t = f(S^Q_t)$. 

\vspace{3pt}
\end{compactenum}

Each of these components and their relation to each other are shown on the left side of \figref{fig:models}. 
We collectively refer to the parameters of the three LSTM models as $\theta_Q$ and 
those of the feature regression network as $\theta_f$. 

\textbf{\Abot} has a similar structure to \Qbot with slight differences since  
it also models the image $I$ via a CNN: 

\begin{compactenum}[\hspace{0pt}-]

\vspace{3pt}
\item \textbf{Question Encoder}: 
\Abot receives a question $q_t$ from \Qbot and encodes it via an LSTM 
$Q^A_t \in \RF^{512}$. 

\vspace{4pt}
\item \textbf{Fact Encoder}: 
Similar to \Qbot, \Abot also encodes the $(q_t, a_t)$-pairs via an LSTM to get $F^A_t \in \RF^{512}$.
The purpose of this encoder is for \Abot to remember what it has already told \Qbot and be able to 
understand references to entities already mentioned. 

\vspace{4pt}
\item \textbf{State/History Encoder} is an LSTM that takes as input at each round $t$ -- 
the encoded question $Q^A_t$, the image features from VGG~\cite{simonyan_iclr15} $y$, 
and the previous fact encoding $F^A_{t-1}$ -- 
to produce a state encoding, \ie 
$\Big( (y, Q^A_1, F^A_{0}), \ldots, (y, Q^A_t, F^A_{t-1}) \Big) \rightarrow S^A_t$. 
This allows the model to contextualize the current question \wrt the history 
while looking at the image to seek an answer.

\vspace{4pt}
\item \textbf{Answer Decoder} is an LSTM that takes the state encoding $S^A_t$ 
and generates $a_t$ by sequentially sampling words. 

\vspace{3pt}
\end{compactenum}

Our code will be publicly available. 

To recap, a dialog round at time $t$ consists of 
1) \Qbot generating a question $q_t$ conditioned on its state encoding $S^Q_{t-1}$,
2) \Abot encoding $q_t$, updating its state encoding $S^A_t$, and generating an answer $a_t$, 
3) \Qbot and \Abot both encoding the completed exchange as $F^Q_t$ and $F^A_t$, and 
4) \Qbot updating its state to $S^Q_t$ based on $F^Q_t$ and making an image representation prediction $\haty_t$ 
for the unseen image. 

\subsection{Joint Training with Policy Gradients}
\label{sec:joint}
In order to train these agents, we use the REINFORCE~\cite{williams1992simple} algorithm 
that updates policy parameters $(\theta_Q, \theta_A, \theta_f)$ in response to experienced rewards.
In this section, we derive the expressions for the parameter gradients for our setup. 

Recall that our agents take actions -- communication $(q_t, a_t)$ and feature prediction $\haty_t$ -- and our 
objective is to maximize the expected reward under the agents' policies, summed over the entire dialog: \\[-25pt]

\begin{center}
\begin{align}
\min_{\theta_A, \theta_Q, \theta_g} 
J(\theta_A, \theta_Q, \theta_g) \quad \text{where,} \\
J(\theta_A, \theta_Q, \theta_g) = 
\mathop{\mathbb{E}}_{\pi_Q, \pi_A} \left[  \sum_{t=1}^T \, r_t \big(s^Q_t, (q_t, a_t, y_t) \big)  \right]
\end{align}
\end{center}
%

\noindent While the above is a natural objective, we find that 
considering the entire dialog as a single RL \emph{episode} does not differentiate between individual good 
or bad exchanges within it. Thus, we update our model based on per-round rewards,
%
\begin{center}
{
\begin{minipage}{\columnwidth}
{\small
\begin{equation}
J(\theta_A, \theta_Q, \theta_g) = \mathop{\mathbb{E}}_{\pi_Q, \pi_A} \left[ r_t \big(s^Q_t, (q_t, a_t, y_t) \big) \right]
\end{equation}
}\end{minipage}}
\end{center}
%
\noindent Following the REINFORCE algorithm, we can write the gradient of this expectation as an 
expectation of a quantity related to the gradient. For $\theta_Q$, we derive this explicitly: 
%

\resizebox{0.85\columnwidth}{!}{
\centering
\begin{minipage}{1.0\columnwidth}
\begin{eqnarray}
\nabla_{\theta_Q}J 
&=& \nabla_{\theta_Q} \left[\mathop{\mathbb{E}}_{\pi_Q, \pi_A} \left[ r_t\left(\cdot \right)  \right] \right] 
\quad \text{($r_t$ inputs hidden to avoid clutter)}
\nonumber \\
&=& \nabla_{\theta_Q} \left[ \sum_{q_t,a_t}\pi_Q\left(q_t | s^Q_{t-1}\right)\pi_A\left(a_t | s^A_{t}\right) r_t\left(\cdot\right)  \right] \nonumber \\
&=& \sum_{q_t,a_t}\pi_Q\left(q_t | s^Q_{t-1}\right)\nabla_{\theta_Q}\log\pi_Q\left(q_t | s^Q_{t-1}\right)\pi_A\left(a_t | s^A_{t}\right) r_t\left(\cdot\right) \nonumber \\
&=& \mathop{\mathbb{E}}_{\pi_Q, \pi_A}\left[ r_t\left(\cdot\right) \, \nabla_{\theta_Q}\log\pi_Q\left(q_t | s^Q_{t-1}\right)  \right]
\hspace{-45pt}
\end{eqnarray}
\end{minipage}}\\[5pt]

\begin{figure*}[t]
	\centering
	\includegraphics[width=\textwidth]{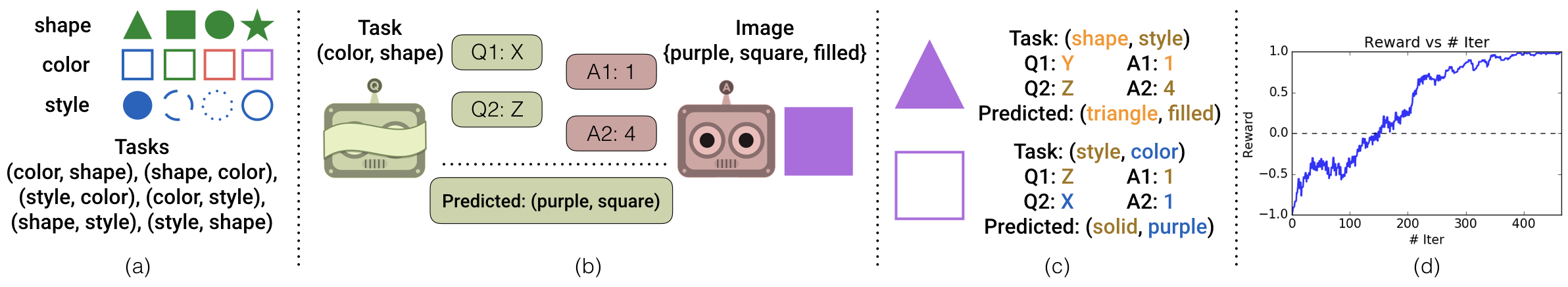}
	\caption{Emergence of grounded dialog: (a) Each `image' has three attributes, 
    and there are six tasks for \Qbot (ordered pairs of attributes). 
    (b) Both agents interact for two rounds followed by attribute pair prediction by \Qbot. 
    (c) Example 2-round dialog where grounding emerges: 
    \emph{color}, \emph{shape}, \emph{style} have been encoded as $X, Y, Z$ respectively.
    (d) Improvement in reward while policy learning.}
	\label{fig:toy-example-qual}
\end{figure*}

Similarly, gradient \wrt $\theta_A$, \ie, $\nabla_{\theta_A}J$ can be derived as
{
\begin{align}
\nabla_{\theta_A} J = \mathop{\mathbb{E}}_{\pi_Q, \pi_A}\left[ r_t\left(\cdot\right) \, \nabla_{\theta_A}\log\pi_A\left(a_t | s^A_{t}\right) \right].
\end{align}
}
As is standard practice, we estimate these expectations with sample averages. Specifically, 
we sample a question from \Qbot (by sequentially sampling words from the question decoder LSTM 
till a stop token is produced), sample its answer from \Abot, compute the scalar reward for this round, 
multiply that scalar reward to gradient of log-probability of this exchange, propagate backward to compute 
gradients \wrt all parameters $\theta_Q, \theta_A$. 
This update has an intuitive interpretation -- 
if a particular $(q_t, a_t)$ is \emph{informative} (\ie, leads to positive reward), its probabilities will be pushed 
up (positive gradient). Conversely, a poor exchange leading to negative reward will be pushed down (negative gradient). 


Finally, since the feature regression network $f(\cdot)$ forms a deterministic policy, 
its parameters $\theta_f$ receive `supervised' gradient updates for differentiable $\ell(\cdot,\cdot)$.

\section{Emergence of Grounded Dialog}
\label{sec:exp_toy}

To succeed at our image guessing game, \Qbot and \Abot need to accomplish a number of challenging sub-tasks 
-- they must learn a common language (\emph{do you understand what I mean when I say `person'?}) and 
develop mappings between symbols and image representations (\emph{what does `person' look like?}),
\ie,
\Abot must learn to ground language in visual perception to answer questions and \Qbot must learn to \sml{predict} plausible image representations 
-- all in an end-to-end manner from a distant reward function.
%
Before diving in to the full task on real images, we conduct a `sanity check' on a synthetic dataset 
with perfect perception to ask  -- \emph{is this even possible?} 

\textbf{Setup.} As shown in \figref{fig:toy-example-qual}, we consider a synthetic world with `images' 
represented as a triplet of attributes -- 4 \emph{shapes}, 4 \emph{colors}, 4 \emph{styles} -- for a total of 64 unique images. 
\Abot has perfect perception and is given direct access to this representation for an image. 
\Qbot is tasked with deducing two 
attributes of the image in a particular order -- 
\eg, 
if the task is 
\emph{(shape, color)}, \Qbot would need to output \emph{(square, purple)}  
for a \emph{(purple, square, filled)} image seen by \Abot (see \reffig{fig:toy-example-qual}b). 
We form 
all 6 such tasks per image. 

\textbf{Vocabulary.} 
We conducted a series of pilot experiments and found the choice of the vocabulary size 
to be crucial for coaxing non-trivial `non-cheating' behavior to emerge. 
For instance, we found that if the \Abot vocabulary $V_A$ is large enough, say $|V_A| \ge 64$ (\#images), 
the optimal policy learnt simply ignores what \Qbot asks and \Abot conveys the entire image in a single token
(\eg token $1 \equiv $ \emph{(red, square, filled)}). 
As with human communication, an impoverished vocabulary that cannot possibly encode the richness 
of the visual sensor is necessary for non-trivial dialog to emerge. 
To ensure at least 2 rounds of dialog, we restrict each agent to only produce a single symbol 
utterance per round from `minimal' vocabularies 
$V_A=\{1,2,3,4\}$ for \Abot and $V_Q=\{X,Y,Z\}$ for \Qbot. Since $|V_A|^{\text{\#rounds}} < \text{\#images}$, 
a non-trivial dialog is necessary to succeed at the task.



\textbf{Policy Learning.} 
Since the action space is discrete and small,
we instantiate \Qbot and \Abot as fully specified tables of Q-values (state, action, future reward estimate) 
and apply tabular Q-learning with Monte Carlo estimation over $10k$ episodes to learn the policies. 
Updates are done alternately where one bot is frozen while the other is updated. 
During training, we use $\epsilon$-greedy policies~\cite{sutton_mitpress98}, ensuring an action probability of $0.6$ for the 
greedy action and split the remaining probability uniformly across other actions. 
At test time, we default to greedy, deterministic policy obtained from these $\epsilon$-greedy policies. 
The task requires outputting the correct attribute value pair based on the task and image. 
Since there are a total of $4+4+4=12$ unique values across the $3$ attributes, \Qbot's final action selects one of 
$12{\times}12{=}144$ attribute-pairs. We use $+1$ and $-1$ as rewards for right and wrong predictions.

\textbf{Results.} 
\figref{fig:toy-example-qual}d shows the reward achieved by the agents' policies \vs number of 
RL iterations (each with 10k episodes/dialogs). 
We can see that the two quickly learn the optimal policy.
%
 \reffig{fig:toy-example-qual}b,c show some example exchanges between the trained bots. 
We find that the two invent their own communication protocol -- 
\Qbot consistently uses specific symbols to query for specific attributes: 
$X \rightarrow$ \emph{color}, $Y \rightarrow$ \emph{shape}, $Z \rightarrow$ \emph{style}. 
And \Abot consistently responds with specific symbols to indicate the inquired attribute, \eg, 
if \Qbot emits $X$ (asks for \emph{color}), \Abot responds with: 
$1 \rightarrow$ \emph{purple}, $2 \rightarrow$ \emph{green}, $3 \rightarrow$ \emph{blue}, $4 \rightarrow$ \emph{red}. 
Similar mappings exist for responses to other attributes. 
Essentially, we find the \emph{automatic emergence of grounded language and a communication protocol} 
among `visual' dialog agents without any human supervision! 

\section{Experiments}
\label{sec:exp}

\afterpage{\clearpage}
\begin{table*}[p]
    \vspace{-40pt}
	\renewcommand*{\arraystretch}{0.8}
    \setlength{\tabcolsep}{8px}
    \begin{center}\hspace{-5.5pt}
    \resizebox{2.10\columnwidth}{!}{
    \begin{tabular}{ p{2.85cm}  p{4cm}  p{5.3cm}  p{6.5cm}  }
    \toprule
    {\small Image + Caption} & {\small Human-Human dialog~\cite{visdial}} & {\small\slp \Qbot-\Abot dialog} & {\small\rl \Qbot-\Abot dialog} \\
    \cmidrule(r){1-1}\cmidrule(lr){2-2}\cmidrule(l){3-3}\cmidrule(l){4-4}
    \raisebox{0pt}{\includegraphics[width=2.7cm]{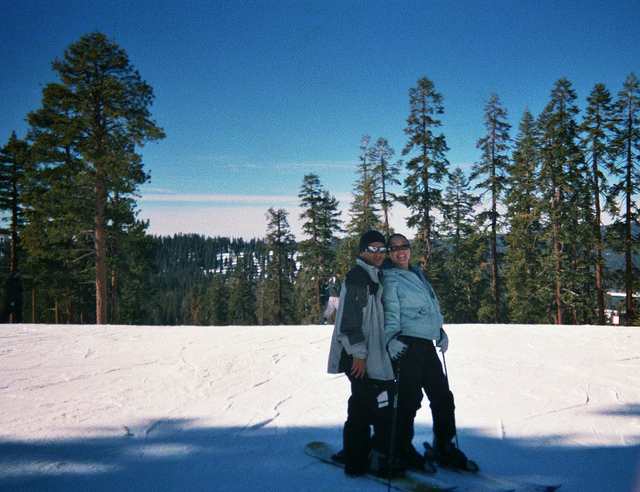}}\newline\vspace{-12pt}
    \begin{minipage}{10cm}{\scriptsize A couple of people in the snow on skis.}\end{minipage}
    &
    \resizebox{4cm}{!}{
    \begin{minipage}{1.55in}
    \vspace{-57pt}
    {\tiny
    \begin{compactenum}[\hspace{-5pt}]
    \item What are their genders? \hfill 1 man 1 woman
    \item Are they both adults? \hfill Yes
    \item Do they wear goggles? \hfill Looks like sunglasses
    \item Do they have hats on? \hfill Man does
    \item Are there any other people? \hfill No
    \item What color is man's hat? \hfill Black
    \item Is it snowing now? \hfill No
    \item What is woman wearing? \hfill Blue jacket and black pants
    \item Are they smiling? \hfill Yes
    \item Do you see trees? \hfill Yes
    \end{compactenum}}
    \end{minipage}}
    &
    \resizebox{5.3cm}{!}{
    \begin{minipage}{2in}
    \vspace{-54pt}
    {\tiny
    \begin{compactenum}[\hspace{-5pt}]
    \item How many people? \hfill 2
    \item Male or female? \hfill  I can't tell
    \item What are they wearing?\hfill I can't tell, they are far away
    \item What color are skis? \hfill I can't tell
    \item Are they wearing goggles? \hfill Yes
    \item Are they wearing goggles? \hfill Yes
    \item Are they wearing goggles? \hfill Yes
    \item Are they wearing goggles? \hfill Yes
    \item Are they wearing hats? \hfill Yes
    \item Are they wearing goggles? \hfill Yes
    \end{compactenum}}
    \end{minipage}}
    &
    \resizebox{6.5cm}{!}{
    \begin{minipage}{2.5in}
    \vspace{-55pt}
    {\tiny
    \begin{compactenum}[\hspace{-5pt}]
    \item Are people male or female or both? \hfill I think 1 is male, can't see other 1 ...
    \item What are they wearing?\hfill They are all wearing snow pants and jackets
    \item Is it snowing? \hfill It does not appear to be snowing at moment
    \item Can you see any trees? \hfill Yes, there are lot of trees in background
    \item Can you see any buildings in background? \hfill No, I can not
    \item Can you see any trees? \hfill Yes, there are lot of trees in background
    \item Can you see any buildings in background? \hfill No, I do not see any buildings
    \item Can you see any trees? \hfill Yes, there are lot of trees in background
    \item Can you see any buildings in background? \hfill No , I do not see any buildings
    \item Does it look like they are in park? \hfill It does not appear to be
    \end{compactenum}}
    \end{minipage}}
    \\[-2pt]  \midrule\iftrue
    \raisebox{0pt}{\includegraphics[width=2.7cm]{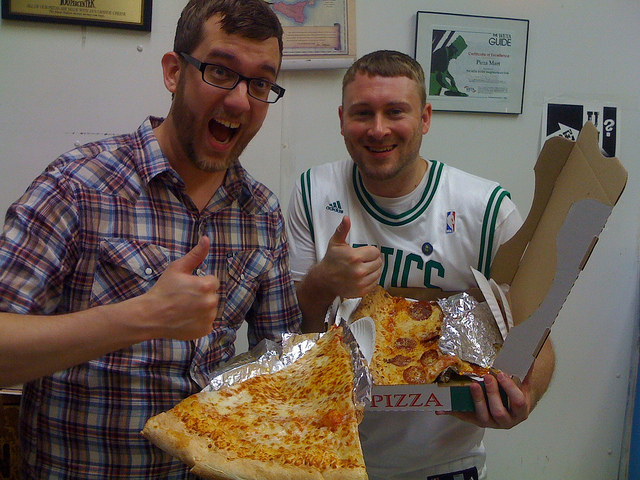}}\newline\vspace{-12pt}
    \begin{minipage}{10cm}{\scriptsize Two men giving thumbs up holding large slices of pizza.}\end{minipage}
    &
    \resizebox{4cm}{!}{
    \begin{minipage}{1.6in}
    \vspace{-54pt}
    {\tiny
    \begin{compactenum}[\hspace{-5pt}]
    \item How old are men? \hfill They look to be 30's
    \item Are they in restaurant? \hfill I can't tell, its close up
    \item Do men look they might be related? \hfill Maybe
    \item Are they each holding slice of pizza in 1 hand and giving thumbs up with other? 1 holding slice, other holding box...
    \item What kind of pizza do they have? \hfill cheese
    \item Does box have pizza company name on it? \hfill No
    \item Are there any pictures or graphics on box?  \hfill No
    \item Are men slender or heavy set? \hfill Slender
    \item Do either of men wear glasses? \hfill Yes
    \item 1 or both? \hfill 1
    \end{compactenum}}
    \end{minipage}}
    &
    \resizebox{5.3cm}{!}{
    \begin{minipage}{2in}
    \vspace{-52pt}
    {\tiny
    \begin{compactenum}[\hspace{-5pt}]
    \item How old are men? \hfill They look like they are in their 20's
    \item Are they male or female? \hfill 1 male and 1 female
    \item What are they wearing? \hfill T shirt and shorts
    \item What kind of pizza is it? \hfill Looks like pepperoni
    \item Does it look good? \hfill Yes it does
    \item Can you tell what kind of pizza it is? \hfill Looks like pepperoni
    \item Are there any toppings on pizza? \hfill Pepperoni
    \item Are there any toppings on pizza? \hfill Pepperoni
    \item Are there any toppings on pizza? \hfill Looks like cheese, but I can't tell...
    \item Are there any other toppings on pizza? \hfill No
    \end{compactenum}}
    \end{minipage}}
    &
    \resizebox{6.5cm}{!}{
    \begin{minipage}{2.6in}
    \vspace{-59pt}
    {\tiny
    \begin{compactenum}[\hspace{-5pt}]
    \item How old are men? \hfill They look like they are in their 20's
    \item Are they wearing chef's hat? \hfill No
    \item Do they have long hair? \hfill No
    \item What kind of pizza is it? \hfill Looks like pepperoni
    \item Does it look good? \hfill Yes
    \item Are there any other people around? \hfill No
    \item Are they inside or outside? \hfill Inside
    \item Can you tell what time of day it is? \hfill No
    \item Are there any other people around? \hfill No
    \item What color are walls? \hfill White
    \end{compactenum}}
    \end{minipage}}
    \\[-2pt] \midrule
    \fi
    \raisebox{0pt}{\includegraphics[width=2.7cm]{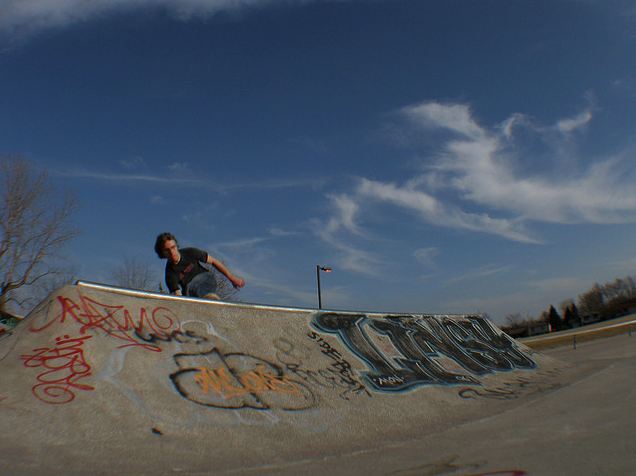}}\newline\vspace{-12pt}
    \begin{minipage}{10cm}{\scriptsize A man riding a skateboard down the side of a ramp.}\end{minipage}
    &
    \resizebox{4cm}{!}{
    \begin{minipage}{1.5in}
    \vspace{-50pt}
    {\tiny
    \begin{compactenum}[\hspace{-5pt}]
    \item How old is man? \hfill In his teens
    \item Is he wearing helmet? \hfill No
    \item Is he wearing any protective gear? \hfill No
    \item What color is his hair? \hfill Dirty blonde or dark brown
    \item What color is his shirt? \hfill Black with red letters
    \item Is he wearing shorts? \hfill Yes
    \item What color are his shorts? \hfill Denim
    \item Are there other people visible? \hfill No
    \item What color are his shoes? \hfill I can't see them
    \item Is this at skate park? \hfill Yes
    \end{compactenum}}
    \end{minipage}}
    &
    \resizebox{5.3cm}{!}{
    \begin{minipage}{2in}
    \vspace{-50pt}
    {\tiny
    \begin{compactenum}[\hspace{-5pt}]
    \item How old is man? \hfill I can't see his face, I don't know
    \item What is he wearing? \hfill shorts and t shirt
    \item What color is his shirt? \hfill white
    \item what color are his pants? \hfill black
    \item what color is his shirt? \hfill blue
    \item What color are his shoes? \hfill Black
    \item What color are his shoes? \hfill Black
    \item What color are his shoes? \hfill Black
    \item What color are his shoes? \hfill Black
    \item What color are his shoes? \hfill Black
    \end{compactenum}}
    \end{minipage}}
    &
    \resizebox{6.5cm}{!}{
    \begin{minipage}{2.5in}
    \vspace{-50pt}
    {\tiny
    \begin{compactenum}[\hspace{-5pt}]
    \item What color is skateboard? \hfill It is hard to tell but I think it's black
    \item Is he wearing helmet? \hfill Yes
    \item Is he wearing knee pads? \hfill No, no knee pads
    \item Is it day or night? \hfill It is daytime
    \item Is it sunny? \hfill It is bright out, but I see no shadows, so I can not
    \item Can you see any other people? \hfill There are 2 other people in background
    \item Is it day or night? \hfill It is daytime
    \item Is it sunny? \hfill Yes, it is sunny
    \item What color is skateboard? \hfill It is hard to tell but I think it's black
    \item Is he wearing helmet? \hfill Yes, he is wearing helmet
    \end{compactenum}}
    \end{minipage}}
    \\[-2pt] \midrule
    \raisebox{0pt}{\includegraphics[width=2.7cm]{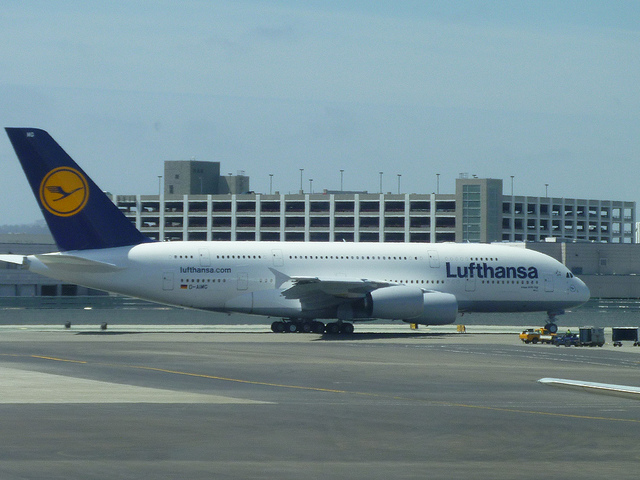}}\newline\vspace{-12pt}
    \begin{minipage}{10cm}{\scriptsize An airplane sitting on an asphalt landing strip.}\end{minipage}
    &
    \resizebox{4cm}{!}{
    \begin{minipage}{1.5in}
    \vspace{-50pt}
    {\tiny
    \begin{compactenum}[\hspace{-5pt}]
    \item What color is plane? \hfill White and blue
    \item Is there writing on plane? \hfill Yes it says lufthansa
    \item What color is ground? \hfill Grey tarmac
    \item Any people? \hfill No people visible
    \item Can you see airport? \hfill Yes
    \item What color is it? \hfill Grey
    \item Is there sunshine? \hfill No it's kind of overcast
    \item Any other planes? \hfill No other planes
    \item Any grass? \hfill No grass
    \item Any trees? \hfill No tree
    \end{compactenum}}
    \end{minipage}}
    &
    \resizebox{5.3cm}{!}{
    \begin{minipage}{2in}
    \vspace{-50pt}
    {\tiny
    \begin{compactenum}[\hspace{-5pt}]
    \item What color is plane? \hfill White
    \item Is it large or small? \hfill Large
    \item Any people? \hfill No
    \item Any writing on plane? \hfill Yes
    \item Any writing on plane? \hfill Yes
    \item Any writing on plane? \hfill Yes
    \item Any writing on plane? \hfill Yes
    \item What does it say? \hfill UNK
    \item What color is plane? \hfill White
    \item Any writing on it? \hfill Yes
    \end{compactenum}}
    \end{minipage}}
    &
    \resizebox{6.5cm}{!}{
    \begin{minipage}{2.5in}
    \vspace{-50pt}
    {\tiny
    \begin{compactenum}[\hspace{-5pt}]
    \item What color is airplane? \hfill White with red and blue stripes
    \item Can you tell what airline it is? \hfill No, it's too far away
    \item What time of day is it? \hfill It is daytime
    \item Can you see any people? \hfill No
    \item Can you see any buildings? \hfill Yes, in background
    \item Is it day or night time? \hfill It is daytime
    \item Can you see sky? \hfill Yes
    \item Are there clouds in sky? \hfill Yes, few
    \item Does it look like it's going to rain? \hfill Yes it does
    \item Does it look like it's going to rain? \hfill Yes it does
    \end{compactenum}}
    \end{minipage}}
    \\[-2pt] \midrule
    \raisebox{0pt}{\includegraphics[width=2.7cm]{figures/main/dog_food.jpg}}\newline\vspace{-12pt}
    \begin{minipage}{10cm}{\scriptsize A dog watching food cook in an oven.}\end{minipage}
    &
    \resizebox{4cm}{!}{
    \begin{minipage}{1.5in}
    \vspace{-50pt}
    {\tiny
    \begin{compactenum}[\hspace{-5pt}]
    \item What color is dog? \hfill Brown and white
    \item What breed is dog? \hfill Look like brittany spaniel
    \item Is dog wearing collar? \hfill Not able to see
    \item Can you tell what is cooking? \hfill Look like beans
    \item Is oven stainless steel? \hfill Yes
    \item Does it have see through glass? \hfill Yes
    \item How many racks in oven? \hfill 2
    \item Can you see knobs? \hfill No
    \item Does oven have wooden handle? \hfill Not able to see
    \item Are beans in pot? \hfill Glass dish
    \end{compactenum}}
    \end{minipage}}
    &
    \resizebox{5.3cm}{!}{
    \begin{minipage}{2in}
    \vspace{-50pt}
    {\tiny
    \begin{compactenum}[\hspace{-5pt}]
    \item What kind of dog is it? \hfill I can't tell
    \item What color is it? \hfill White
    \item Is it large or small? \hfill Large
    \item What kind of food is it? \hfill I can't tell
    \item Is it large or small? \hfill Small
    \item What color is it? \hfill White
    \item Is it large or small? \hfill Large
    \item What color are walls? \hfill White
    \item What kind of floor? \hfill Can't see floor
    \item What kind of floor? \hfill Can't see floor
    \end{compactenum}}
    \end{minipage}}
    &
    \resizebox{6.5cm}{!}{
    \begin{minipage}{2.5in}
    \vspace{-50pt}
    {\tiny
    \begin{compactenum}[\hspace{-5pt}]
    \item What kind of dog is it? \hfill Terrier of some kind
    \item What color is it? \hfill White and grey
    \item Can you tell what kind of food it is? \hfill No
    \item Are there people around? \hfill No people
    \item Is it inside or outside? \hfill Inside
    \item Can you tell what time of day it is? \hfill No I can't
    \item What color is room? \hfill I can't see walls
    \item Do you see any windows? \hfill No
    \item Do you see any windows? \hfill No
    \item Do you see any windows? \hfill No
    \end{compactenum}}
    \end{minipage}}
    \\[-2pt] \midrule
    \raisebox{0pt}{\includegraphics[width=2cm,height=2cm]{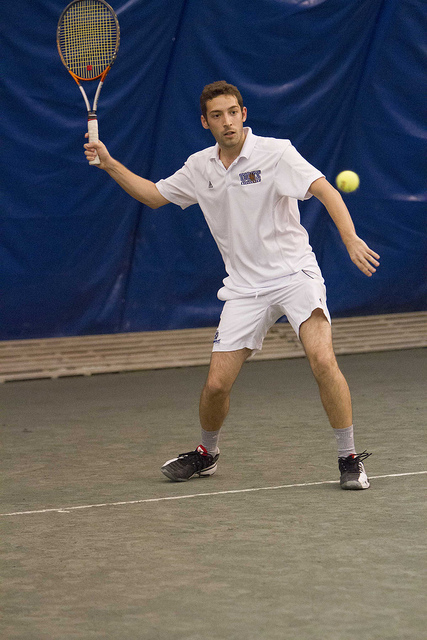}}\newline\vspace{-12pt}
    \begin{minipage}{10cm}{\scriptsize A man prepares to swing at the tennis ball.}\end{minipage}
    &
    \resizebox{4cm}{!}{
    \begin{minipage}{1.5in}
    \vspace{-50pt}
    {\tiny
    \begin{compactenum}[\hspace{-5pt}]
    \item Are there more people in this photo or just man? \hfill No
    \item Just tennis player man in this photo? \hfill Yes
    \item Is it outdoor court? \hfill Can't tell, it's close up
    \item What is man wearing? \hfill White shirt and shorts
    \item Can you see net? \hfill No
    \item Can you see bottom of court? \hfill Yes
    \item Can you see ball in photo? \hfill Yes
    \item Is it close to man's racquet? \hfill Yes
    \item Is man about to take swing? \hfill Yes
    \item Is he left handed or right handed? \hfill Right
    \end{compactenum}}
    \end{minipage}}
    &
    \resizebox{5.3cm}{!}{
    \begin{minipage}{2in}
    \vspace{-50pt}
    {\tiny
    \begin{compactenum}[\hspace{-5pt}]
    \item How old is man? \hfill I can't see his face, I don't know
    \item What color is his hair? \hfill Brown
    \item What is he wearing? \hfill Shorts and t shirt
    \item What color is his shirt? \hfill White
    \item What color is his shirt? \hfill White
    \item What color is his shirt? \hfill White
    \item What color is his shirt? \hfill White
    \item What color is his shirt? \hfill White
    \item What color is his shirt? \hfill White
    \item What color is his shirt? \hfill White
    \end{compactenum}}
    \end{minipage}}
    &
    \resizebox{6.5cm}{!}{
    \begin{minipage}{2.5in}
    \vspace{-50pt}
    {\tiny
    \begin{compactenum}[\hspace{-5pt}]
    \item How old is man? \hfill He looks to be in his 20's
    \item What color is his hair? \hfill His hair is brown
    \item Is he wearing glasses? \hfill No he is not
    \item Is he wearing hat? \hfill No he is not
    \item Is he wearing glasses? \hfill No he is not
    \item Are there other people around? \hfill No it's just him
    \item Can you see net? \hfill No I can't
    \item What color is court? \hfill Court is green
    \item Is it sunny? \hfill Yes it is
    \item Do you see any trees? \hfill No I do not
    \end{compactenum}}
    \end{minipage}}
    \\[-2pt] \midrule
    \raisebox{0pt}{\includegraphics[height=2.3cm,width=2.7cm]{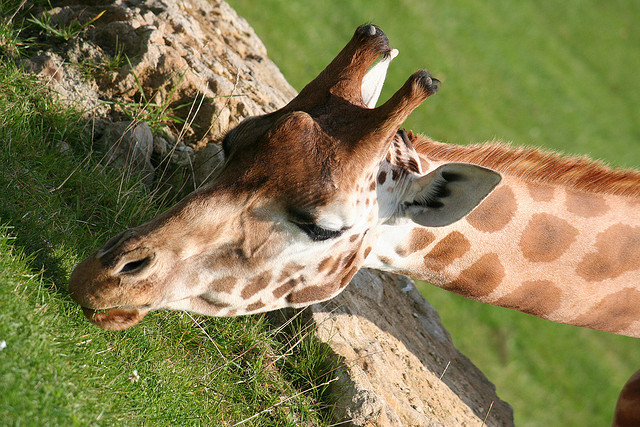}}\newline\vspace{-12pt}
    \begin{minipage}{10cm}{\scriptsize A giraffe is resting its head in the grass near a rock.}\end{minipage}
    &
    \resizebox{4cm}{!}{
    \begin{minipage}{1.5in}
    \vspace{-63pt}
    {\tiny
    \begin{compactenum}[\hspace{-5pt}]
    \item Is giraffe adult? \hfill I cannot tell, I only see face
    \item Is grass at higher altitude than its feet? \hfill No
    \item Is rock big? \hfill Yes
    \item How is weather? \hfill Sunny
    \item What color is rock? \hfill Tan color
    \item Is grass green or dry? \hfill Green but looks little dry
    \item Does giraffe look healthy? \hfill Yes
    \item Does giraffe look happy? \hfill It looks tired
    \item Is giraffe laying down? \hfill Yes
    \item Is giraffe's mouth closed? \hfill Yes
    \end{compactenum}}
    \end{minipage}}
    &
    \resizebox{5.3cm}{!}{
    \begin{minipage}{2in}
    \vspace{-65pt}
    {\tiny
    \begin{compactenum}[\hspace{-5pt}]
    \item Is there only 1 giraffe? \hfill Yes
    \item Are there any other animals? \hfill No
    \item Are there any people? \hfill No
    \item Are there any other animals? \hfill No
    \item Is it day or night? \hfill Day
    \item Is it sunny? \hfill Yes
    \item What time of day it is? \hfill I can't tell
    \item Are there any other animals? \hfill No
    \item Are there any trees? \hfill No
    \item Are there any other animals? \hfill No
    \end{compactenum}}
    \end{minipage}}
    &
    \resizebox{6.5cm}{!}{
    \begin{minipage}{2.5in}
    \vspace{-65pt}
    {\tiny
    \begin{compactenum}[\hspace{-5pt}]
    \item Is there only 1 giraffe? \hfill Yes
    \item Are there any other animals? \hfill No
    \item Are there any people? \hfill No
    \item Does this look like zoo? \hfill Can't tell
    \item Is there fence? \hfill No
    \item Is it day or night? \hfill Day
    \item Is it sunny? \hfill Yes
    \item What time of day is it? \hfill Morning
    \item Are there any trees? \hfill No
    \item Can you see sky? \hfill No
    \end{compactenum}}
    \end{minipage}}
    \\[-2pt] \midrule
    \raisebox{0pt}{\includegraphics[height=2.6cm,width=2.7cm]{figures/main/cracker.jpg}}\newline\vspace{-12pt}
    \begin{minipage}{10cm}{\scriptsize A cracker with shredded cheese and leafy greens on a white plate.}\end{minipage}
    &
    \resizebox{4cm}{!}{
    \begin{minipage}{1.5in}
    \vspace{-63pt}
    {\tiny
    \begin{compactenum}[\hspace{-5pt}]
    \item Are there any people in picture? \hfill No, there are no people
    \item Is cheese white? \hfill Yes, cheese is white
    \item Are crackers white? \hfill No, cracker is brownish
    \item Are there lots of greens? \hfill No, only 1 sprig
    \item Was picture taken in kitchen? \hfill I cannot tell, it is too zoomed in
    \item How many crackers are there? \hfill There is 1
    \item Is there design on plate? \hfill No, plate is plane white
    \item Is there good lighting? \hfill Yes, lighting is decent
    \item Is plate on table? \hfill I cannot tell I can only see plate as background
    \item Does cracker look delicious? \hfill Absolutely
    \end{compactenum}}
    \end{minipage}}
    &
    \resizebox{5.3cm}{!}{
    \begin{minipage}{2in}
    \vspace{-83pt}
    {\tiny
    \begin{compactenum}[\hspace{-5pt}]
    \item Are there any people? \hfill No
    \item What color is plate? \hfill White
    \item Is plate on table? \hfill I can't tell
    \item What color is plate? \hfill White
    \item Is plate on table? \hfill I can't tell
    \item Is there anything else on table? \hfill No
    \item Is there anything else on table? \hfill No
    \item Is there anything else on table? \hfill No
    \item Is there anything else on table? \hfill No
    \item Is there anything else on table? \hfill No
    \end{compactenum}}
    \end{minipage}}
    &
    \resizebox{6.5cm}{!}{
    \begin{minipage}{2.5in}
    \vspace{-83pt}
    {\tiny
    \begin{compactenum}[\hspace{-5pt}]
    \item Is plate large or small? \hfill Large
    \item Is it on table? \hfill I can't tell
    \item What kind of donut? \hfill Plain with sugar on it
    \item What kind of chips? \hfill Potato chips
    \item Any people? \hfill No
    \item Is this inside or outside? \hfill I can't tell, it's close up
    \item What color is table? \hfill White
    \item Any chairs? \hfill No
    \item What color is table? \hfill I can't tell
    \item Is it large or small? \hfill I can't see plate
    \end{compactenum}}
    \end{minipage}}
    \\[-2pt] \midrule
    \raisebox{0pt}{\includegraphics[height=2.6cm,width=2.3cm]{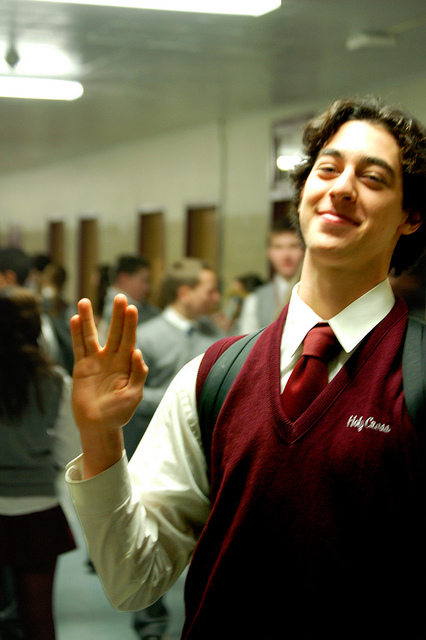}}\newline\vspace{-12pt}
    \begin{minipage}{10cm}{\scriptsize A man making the live long and prosper sign from star trek.}\end{minipage}
    &
    \resizebox{4cm}{!}{
    \begin{minipage}{1.5in}
    \vspace{-63pt}
    {\tiny
    \begin{compactenum}[\hspace{-5pt}]
    \item How old is this man? \hfill He looks like late teens
    \item Does he have short hair or long hair? \hfill Medium
    \item What color is his hair? \hfill Brown
    \item Is he wearing glasses? \hfill No
    \item Does he have facial hair? \hfill No
    \item How is he dressed? \hfill He's white dress shirt with red tie and matching red vest over it looks like school uniform
    \item Is he larger man or is he skinny? \hfill He's skinny
    \item Is this picture taken outside? \hfill Inside
    \item Can you tell if it's daytime? \hfill I'm not sure
    \item Can you tell what room he is standing in? \hfill He's in hallway
    \end{compactenum}}
    \end{minipage}}
    &
    \resizebox{5.3cm}{!}{
    \begin{minipage}{2in}
    \vspace{-83pt}
    {\tiny
    \begin{compactenum}[\hspace{-5pt}]
    \item How old is man? \hfill He looks like he's in his 20s
    \item What is he wearing? \hfill T shirt and jeans
    \item What color is his hair? \hfill Black
    \item What is he wearing? \hfill T shirt and jeans
    \item What color is his hair? \hfill Black
    \item What is he wearing? \hfill T shirt and jeans
    \item What color is his hair? \hfill Black
    \item Is it short or long? \hfill Short
    \item What color is his hair? \hfill Black
    \item Is it short or long? \hfill Short
    \end{compactenum}}
    \end{minipage}}
    &
    \resizebox{6.5cm}{!}{
    \begin{minipage}{2.5in}
    \vspace{-80pt}
    {\tiny
    \begin{compactenum}[\hspace{-5pt}]
    \item Where is man located? \hfill Looks like classroom of some sort
    \item What is man wearing? \hfill Black t shirt and jeans
    \item Are there any other people in picture? \hfill Yes, quite few people in background
    \item What color is table? \hfill Picture is black and white, but it's wood table
    \item Is there anything else on table? \hfill Not that I can see
    \item What are other people doing? \hfill They are standing in front of him, but I don't know what they
    \item Are there any windows? \hfill Not that I can see
    \item What color are walls? \hfill I can't see walls
    \item What time of day do you think it is? \hfill I can't tell
    \item What is man wearing? \hfill Black t shirt and jeans
    \end{compactenum}}
    \end{minipage}}
    \\ \bottomrule
    \end{tabular}}\\[5pt]
    \caption{Selected examples of \Qbot-\Abot interactions for \slp~and \rl.
    \rl~interactions are diverse, less prone to repetitive and safe exchanges (``can't tell", ``don't know"~\etc), and more image-discriminative.}
    \label{table:sl_rl}
    \end{center}
\end{table*}

Our synthetic experiments in the previous section establish that when faced with a cooperative task where information
must be exchanged, two agents with perfect perception are capable of developing a complex communication protocol. 

In general, with imperfect perception on real images, discovering human-interpretable language
and communication strategy from scratch
is both tremendously difficult and an unnecessary re-invention of English.
We leverage the recently introduced
VisDial dataset~\cite{visdial} that contains (as of the publicly released v0.5) human dialogs
(10 rounds of question-answer pairs) on 68k images from the COCO dataset, for a total of 680k QA-pairs.
Example dialogs from the VisDial dataset are shown in \reftab{table:sl_rl}.





\textbf{Image Feature Regression.}
We consider a specific instantiation of the visual guessing game described in \refsec{sec:game} --
specifically at each round $t$, \Qbot needs to regress to the vector embedding $\haty_t$
of image $I$ corresponding to the fc7 (penultimate fully-connected layer)
output from VGG-16~\cite{simonyan_iclr15}. The distance metric used in the reward computation is
$\ell_2$, \ie
$r_t ( \cdot ) = \norm{y^{gt}-\haty_{t-1}}^2_2
- \norm{y^{gt} - \haty_t}^2_2$.



\textbf{Training Strategies.}
We found two training strategies to be crucial to ensure/improve the convergence of the RL framework described
in \refsec{sec:model}, to produce any meaningful dialog exchanges, and to ground the agents in natural language.
\begin{asparaenum}[\hspace{0pt}1)]
\item \textbf{Supervised Pretraining.}
We first train both agents in a supervised manner
on the \train split of VisDial~\cite{visdial} v0.5 under an MLE objective.
Thus, conditioned on human dialog history, \Qbot is trained to generate the follow-up question by human1,
\Abot is trained to generate the response by human2,  
and the feature network $f(\cdot)$ is optimized to regress to $y$.
The CNN in \Abot is pretrained on ImageNet.
This pretraining ensures that the agents can generally recognize some objects/scenes
and emit English questions/answers. The space of possible
$(q_t, a_t)$ is tremendously large and without pretraining
most exchanges result in no information gain about the image.

\item \textbf{Curriculum Learning.}
After supervised pretraining, we `smoothly' transition the agents to RL training
according to a curriculum.
Specifically, we continue supervised training for the first $K$ (say 9) rounds of dialog and transition to
policy-gradient updates for the remaining $10 - K$ rounds. We start at $K=9$ and
gradually anneal to 0.
This curriculum ensures that the agent team does not suddenly diverge off policy,
if one incorrect $q$ or $a$ is generated. 

\end{asparaenum}

Models are pretrained for 15 epochs on VisDial, after which we transition to policy-gradient training
by annealing $K$ down by $1$ every epoch.
All LSTMs are $2$-layered with $512$-d hidden states.
We use Adam~\cite{kingma_iclr15} with a learning rate of $10^{-3}$, and clamp gradients
to $[-5, 5]$ to avoid explosion.
All our code will be made publicly available.
There is no explicit state-dependent baseline in our training as we initialize from supervised pretraining
and have zero-centered reward, which ensures a good proportion of random samples are
both positively and negatively reinforced.

\textbf{Model Ablations.}
We compare to a few natural ablations of our full model, denoted \rl.
First, we evaluate the purely supervised agents (denoted \slp),
\ie, trained only on VisDial data (no RL).
Comparison to these agents establishes how much RL helps over supervised learning.
Second, we fix one of \Qbot or \Abot to the supervised pretrained initialization
and train the other agent (and the regression network $f$) with RL;
we label these as \fq~or \fa~respectively.
Comparing to these partially frozen agents tell us the importance
of coordinated communication.
Finally, we freeze the regression network $f$ to the supervised pretrained
initialization while training \Qbot and \Abot with RL.
This measures improvements from language adaptation alone.

We quantify performance of these agents along two dimensions --
how well they perform on the image guessing task (\ie image retrieval)
and how closely they emulate human dialogs
(\ie performance on VisDial dataset~\cite{visdial}).

\begin{figure*}[th]
	\renewcommand*{\arraystretch}{0.9}
    \setlength{\tabcolsep}{1pt}
    \centering
    \resizebox{2\columnwidth}{!}{
    \begin{tabular}[t]{ c c }
        \resizebox{0.50\textwidth}{!}{
        \smallskip
        \begin{subfigure}{0.45\textwidth}
            \centering
            \includegraphics[clip=true, width=\textwidth]{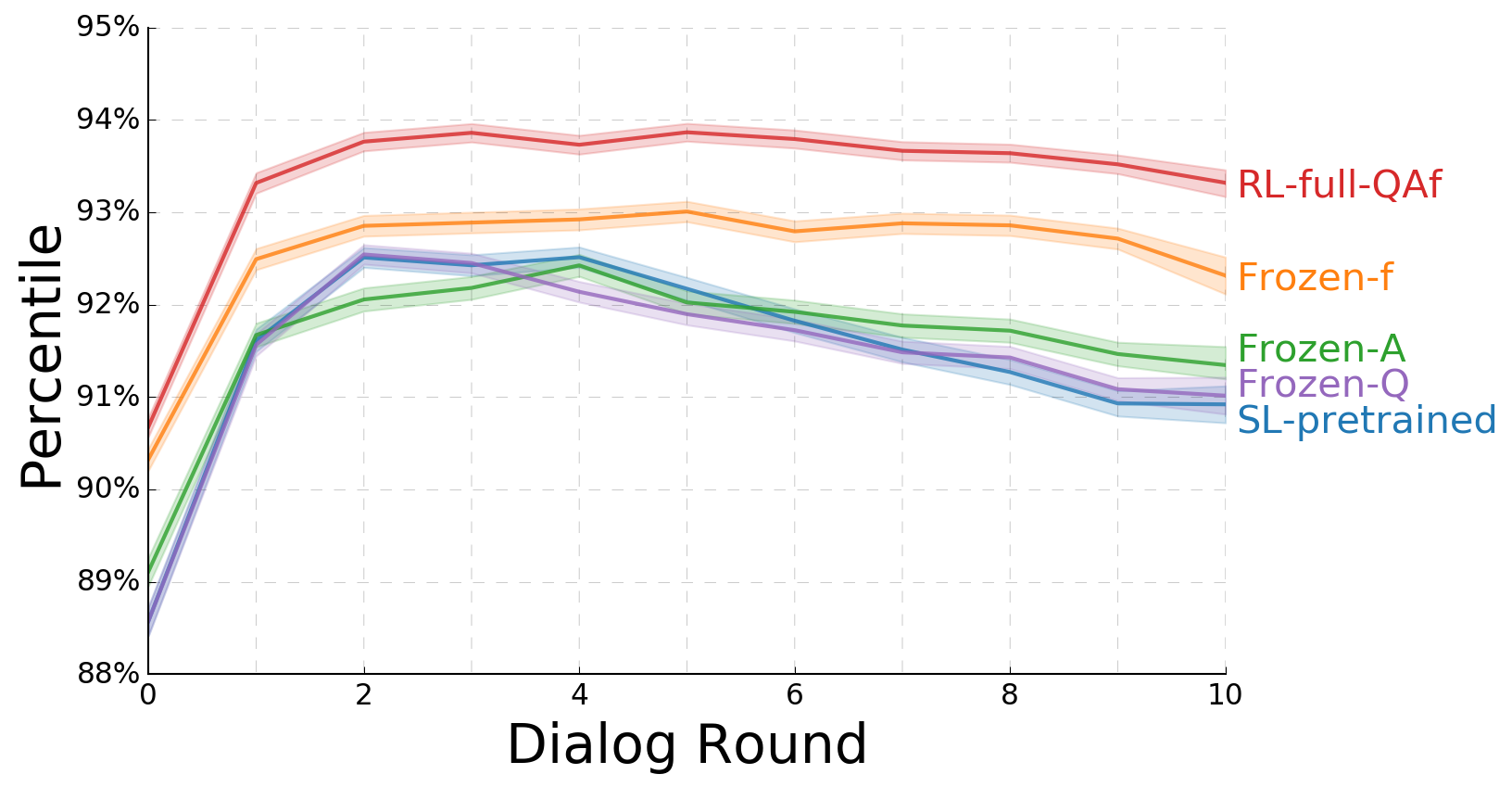}
            \caption{Guessing Game Evaluation.}
            \label{fig:image_eval}
        \end{subfigure}} &
        \resizebox{0.5\textwidth}{!}{
        \begin{subfigure}{0.45\textwidth}
            \centering
            \vspace{30pt}
            \resizebox{0.9\textwidth}{!}{
            \setlength{\tabcolsep}{5pt}
            \begin{tabular}{cccccc}
            \toprule
             \textbf{Model} & \textbf{MRR} & \textbf{R@5} & \textbf{R@10} & \textbf{Mean Rank}\\
            \midrule
            SL-pretrain & 0.436 & 53.41 & 60.09 & 21.83 \\[1.5pt]
            Frozen-Q & 0.428 & 53.12 & 60.19 & 21.52 \\[1.5pt]
            Frozen-f & 0.432 & 53.28 & 60.11 & 21.54 \\[1.5pt]
            RL-full-QAf & 0.428 & 53.08 & 60.22 & 21.54 \\[1.5pt]
            Frozen-Q-multi & \textbf{0.437} & \textbf{53.67} & \textbf{60.48} & \textbf{21.13} \\[1.5pt]
            \bottomrule
            \end{tabular}}\\[15pt]
            \caption{Visual Dialog Answerer Evaluation.}
            \label{table:qa_eval}
        \end{subfigure}}
    \end{tabular}}\\[5pt]
    \begin{subfigure}{\textwidth}
        \centering
        \includegraphics[clip=true, width=\textwidth]{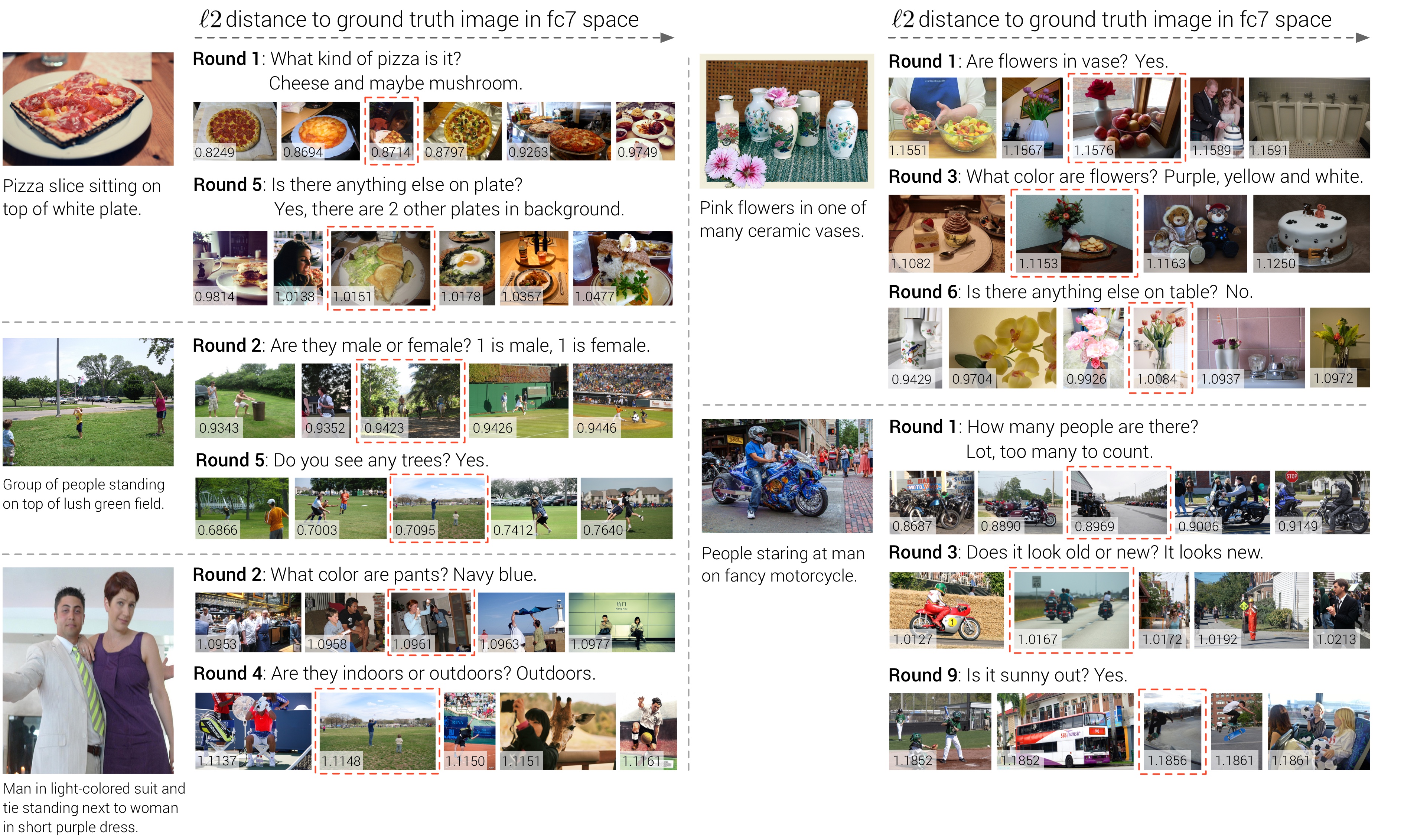}
        \vspace{-20pt}
        \caption{Qualitative Retrieval Results.}
        \label{fig:nn_examples}
    \end{subfigure}
    \caption{\textbf{a) Guessing Game Evaluation.}
    Plot shows the rank in percentile (higher is better) of the `ground truth' image (shown to \Abot)
    as retrieved using fc7 predictions of \Qbot \vs rounds of dialog.
    Round 0 corresponds to image guessing based on the caption alone.
    We can see that the \rl bots significantly outperforms the \slp bots (and other ablations).
    Error bars show standard error of means.
    \textbf{(c)} shows qualitative results on this predicted fc7-based image retrieval.
    Left column shows true image and caption, right column shows dialog exchange, and
    a list of images sorted by their distance to the ground-truth image. The image predicted by \Qbot
    is highlighted in red. We can see that the predicted image is often semantically quite similar.
    \textbf{b) VisDial Evaluation.} Performance of \Abot on VisDial v0.5 \test,
    under mean reciprocal rank (MRR), recall@$k$ for $k=\{5, 10\}$ and mean rank metrics.
    Higher is better for MRR and recall@$k$, while lower is better for mean rank.
    We see that our proposed \fqm~outperforms all other models on VisDial metrics by 3\% relative gain.
    This improvement is entirely `for free' since no additional annotations were required for RL.}
\end{figure*}
\textbf{Evaluation: Guessing Game.}
To assess how well the agents have learned to cooperate at the image
guessing task, we setup an image retrieval experiment based on the
\test split of VisDial v0.5 ($\sim$$9.5k$ images), which were
never seen by the agents in RL training.
We present each image + an automatically generated caption~\cite{karpathy_cvpr15}
to the agents, and allow them to communicate over 10 rounds of dialog.
After each round, \Qbot predicts a feature representation $\haty_t$.
We sort the entire \test set in ascending distance to this prediction
and compute the rank of the source image.

\reffig{fig:image_eval} shows the mean percentile rank of the source
image for our method and the baselines across the rounds (shaded region indicates standard error).
A percentile rank of 95\% means that the source image is closer to the prediction than 95\%
of the images in the set. \reftab{table:sl_rl} shows example exchanges
between two humans (from VisDial), the \slp~
and the \rl~agents.
We make a few observations:
%
\begin{itemize}

\item \textbf{RL improves image identification.}
We see that \rl~significantly outperforms \slp~
and all other ablations (\eg, at round 10, improving percentile rank
by over $3\%$), indicating that our training
framework is indeed effective at training these agents for image guessing.

\item \textbf{All agents `forget'; RL agents forget less.}
One interesting trend we note in \reffig{fig:image_eval} is that
all methods significantly improve from round $0$
(caption-based retrieval) to rounds 2 or 3, but beyond that
all methods with the exception of \rl~get \emph{worse},
even though they have strictly more information. 
As shown in \reftab{table:sl_rl}, 
agents will often get
stuck in infinite repeating loops
but this is much rarer for RL agents. Moreover, even when RL agents repeat
themselves, it is after longer gaps (2-5 rounds).
We conjecture that the goal of helping a partner over multiple  rounds encourages longer term memory retention.

\item \textbf{RL leads to more informative dialog.}
SL \Abot tends to produce `safe' generic responses
(\myquote{I don't know}, \myquote{I can't see}) but RL \Abot responses
are much more detailed (\myquote{It is hard to tell but I think it's black}).
These observations are consistent with recent literature in text-only
dialog~\cite{li_emnlp16}.
Our hypothesis for this improvement is that
human responses are diverse and SL trained agents tend to `hedge their bets'
and achieve a reasonable log-likelihood by being non-committal. In contrast,
such `safe' responses do not help \Qbot in picking the correct image, thus
encouraging an informative RL \Abot.

\end{itemize}


\textbf{Evaluation: Emulating Human Dialogs.}
To quantify how well the agents emulate human dialog,
we evaluate \Abot on the retrieval metrics proposed by Das~\etal~\cite{visdial}.
Specifically, every question in VisDial is
accompanied by 100 candidate responses. We use the log-likehood assigned
by the \Abot answer decoder to sort these candidates and report
the results in \reftab{table:qa_eval}. We find that despite the
RL \Abot's answer being more informative,
the improvements on VisDial metrics are minor.
We believe this is
because while the answers are correct, they may not necessarily mimic
human responses (which is what the answer retrieval metrics check for).
In order to dig deeper,
we train a variant of \fq~with a multi-task objective
-- simultaneous (1) ground truth answer supervision
and (2) image guessing reward, to keep \Abot close to human-like
responses.
We use a weight of 1.0 for the SL loss and 10.0 for RL.
This model, denoted \fqm,
performs better than all other approaches 
on VisDial answering metrics, improving the best reported result
on VisDial by 0.7 mean rank (relative improvement of $3\%$).
Note that this gain is entirely `free' since no additional annotations were required for RL.

\textbf{Human Study.}
We conducted a human interpretability study to measure
(1) whether humans can easily understand the \Qbot-\Abot dialog, and
(2) how image-discriminative the interactions are.
We show human subjects a pool of 16 images, the agent dialog (10 rounds),
and ask humans to pick their top-5 guesses for the image the two agents
are talking about.
We find that mean rank of the ground-truth image for \slp~
agent dialog is 3.70 \vs 2.73 for \rl~dialog.
In terms of MRR, the comparison is 0.518 \vs 0.622 respectively.
Thus, under both metrics, humans find it easier to guess the unseen image based on
\rl~dialog exchanges, which shows that
agents trained within our framework
(1) successfully develop image-discriminative language, and
(2) this language is interpretable; they do not deviate off English.



\section{Conclusions}
\label{sec:conclusions}

To summarize, 
we introduce a novel training framework for visually-grounded dialog agents by posing 
a cooperative `image guessing' game between two agents. 
%
%
%
We use deep reinforcement learning to learn the policies of these agents  
end-to-end -- from pixels to multi-agent multi-round dialog to game reward. 
%
%
We demonstrate the power of this framework in a completely ungrounded 
synthetic world, where the agents communicate via symbols 
with no pre-specified meanings (X, Y, Z). 
We find that two bots invent their own communication protocol 
without any human supervision.  
We go on to instantiate this game on the VisDial~\cite{visdial} dataset, where 
we pretrain with supervised dialog data. We find that the RL `fine-tuned' agents not only significantly 
outperform SL agents, but learn to play to each other's strengths, 
all the while 
remaining interpretable to outside humans observers. 

\paragraph{Acknowledgements.}
We thank Devi Parikh for helpful discussions. 
This work was funded in part by the following awards to DB --
NSF CAREER award,
ONR YIP award, ONR Grant N00014-14-1-0679,
ARO YIP award,
ICTAS Junior Faculty award,
Google Faculty Research Award,
Amazon Academic Research Award,
AWS Cloud Credits for Research, and NVIDIA GPU donations.
SK was supported by ONR Grant N00014-12-1-0903,
and SL was partially supported by the Bradley Postdoctoral Fellowship.
Views and conclusions contained herein are those of the authors and should not be interpreted as necessarily representing the
official policies or endorsements, either expressed or implied, of the U.S. Government, or any sponsor.

{
\small
\bibliographystyle{ieee}
\bibliography{strings,main}
}

\end{document}